\pdfoutput=1
\documentclass[10pt,letterpaper]{article}

\usepackage[english]{babel}
\usepackage[utf8]{inputenc}
\usepackage[T1]{fontenc}

\usepackage{booktabs}
\usepackage{array}
\usepackage{multirow}
\newcolumntype{L}[1]{>{\raggedright\arraybackslash}p{#1}}

\usepackage[letterpaper,left=1.5in,right=1.5in,top=1in,bottom=1in]{geometry}

\usepackage{amsmath}
\usepackage{amssymb}
\usepackage{newtxtext,newtxmath}
\usepackage{bm}
\usepackage{enumitem}
\usepackage{microtype}

\makeatletter
\newcommand{\plainpercent}{{\usefont{T1}{lmr}{\f@series}{n}\char37}}
\makeatother
\renewcommand{\%}{\plainpercent}

\usepackage[compact]{titlesec}
\titleformat{\section}{\large\bfseries}{\thesection}{0.6em}{}
\titleformat{\subsection}{\normalsize\bfseries}{\thesubsection}{0.6em}{}
\titlespacing{\section}{0pt}{1.2\baselineskip}{0.5\baselineskip}

\usepackage{graphicx}
\graphicspath{{figures/}}
\usepackage{xurl}
\usepackage[colorlinks=true, allcolors=blue]{hyperref}

\hypersetup{
  pdftitle={Rollout-Decoded Reconstruction for Long-Horizon Prediction in Latent World Models},
  pdfauthor={Rishi Shah and Rishav Shrestha},
  pdfkeywords={world models; latent dynamics; decoder; exposure bias; rollout;
               Kuramoto-Sivashinsky; chaotic forecasting; valid prediction time;
               model-based control; model predictive control},
}

\usepackage{float}
\usepackage[font=small,labelfont=bf,labelsep=colon,skip=6pt]{caption}

\newcommand{\kfig}[2]{\IfFileExists{#1}{\includegraphics[width=#2]{#1}}{\fbox{\parbox[c][10em][c]{#2}{\centering\texttt{#1}\\(figure placeholder)}}}}
\newcommand{\sg}{\mathrm{sg}}
\newcommand{\tu}{\,\mathrm{tu}}

\title{Rollout-Decoded Reconstruction for Long-Horizon\\ Prediction in Latent World Models}

\author{%
  \begin{tabular}{c@{\hspace{0.9in}}c}
    \textbf{Rishi Shah}            & \textbf{Rishav Shrestha}          \\
    Machine Learning Engineer      & Chief Technical Officer           \\
    E3A Healthcare                 & E3A Healthcare                    \\
    \texttt{rishishah994@gmail.com} & \texttt{rishav@e3ahealth.com}    \\
  \end{tabular}
}

\makeatletter
\renewcommand{\maketitle}{%
  \null\vskip 6pt
  \begin{center}
    {\rule{\linewidth}{1.4pt}}\vskip 8pt
    {\LARGE\bfseries \@title \par}\vskip 8pt
    {\rule{\linewidth}{0.5pt}}\vskip 22pt
    {\normalsize \@author \par}\vskip 16pt
  \end{center}
  \vskip 20pt
}
\makeatother

\newenvironment{abstractblock}{%
  \begin{center}{\bfseries\normalsize Abstract}\end{center}
  \vspace{-0.4em}
  \begin{list}{}{\setlength{\leftmargin}{0.4in}\setlength{\rightmargin}{0.4in}}
  \item[]%
}{%
  \end{list}\vspace{0.8em}
}

\begin{document}
\maketitle

\begin{abstractblock}
A latent world model trains its decoder on latents anchored to observations, then deploys it on the model's own free-running rollout, hundreds of steps past the last observation. Rollout-Decoded Reconstruction (RDR) closes this gap with a single loss term that free-runs the model during training exactly as evaluation will, decodes every rollout latent, and penalizes reconstruction error against ground truth. The term adds no parameters, costs training-time compute only, and reduces to the standard objective at weight zero, so every comparison in this paper is a one-flag A/B. On the chaotic Kuramoto--Sivashinsky equation, RDR raises valid prediction time (the time to first crossing of normalized error 0.5) from $3.87 \pm 0.23$ to $6.97 \pm 0.42$ time units at an identical 193{,}568 parameters, a 1.80$\times$ improvement confirmed on seeds never used in selection and holding in 10 of 10 preregistered configurations at ratios of 1.71--2.50$\times$. The results come from a single system; a sweep in which the advantage grows with latent width is descriptive, and control experiments on two classic tasks are preliminary.
\end{abstractblock}

\section{Introduction}
\label{sec:intro}

A latent world model predicts by free-running: an encoder compresses the last observation into a latent state, a transition iterates that latent forward on its own outputs, and a decoder turns each rollout latent into a predicted observation~\cite{hafner2019,hafner2020,hafner2021,hafner2025}. The decoder, however, is trained only on latents anchored to observations: encoder outputs, and predictions a single teacher-forced step from them. Nothing in the standard objective ties decoding quality to the drifted latents a long rollout visits, so a decoder that is sharp near observations can fail many steps into the rollout it exists to decode. PlaNet~\cite{hafner2019} named the direct fix, decoding the rollout during training, and set it aside; no latent world model since has made it the object of study (Section~\ref{sec:background}).

We propose Rollout-Decoded Reconstruction (RDR), a loss term that closes this gap directly. During training, the model is rolled free-running from an encoded initial state, exactly as evaluation will roll it; every rollout latent is decoded; and the reconstruction error against ground truth is penalized. The term adds no parameters, changes no architecture, and reduces to the standard objective when its weight $\lambda$ is zero. Every preregistered comparison is therefore an A/B at fixed data, seeds, budget, and parameter count, differing in one flag; the added cost is training-time compute (Appendix~\ref{app:compute}).

On the Kuramoto--Sivashinsky (KS) equation~\cite{kuramoto1976,sivashinsky1977}, a chaotic PDE with exact ground truth and a standard long-horizon metric, RDR raises valid prediction time (VPT, the time to first crossing of normalized error 0.5) from $3.87 \pm 0.23$ to $6.97 \pm 0.42$ time units: a 1.80$\times$ improvement at an identical 193{,}568 parameters, confirmed on fresh seeds and at both evaluation horizons, with the direction holding in 10 of 10 preregistered configurations at ratios of 1.71--2.50$\times$ (Section~\ref{sec:headline}). The effect is consistent with the training-distribution account: a variant that spends +25.7\% more parameters on the same objective performs worse in 7 of 10 configurations, which rules out added capacity as the explanation, and both arms' rollout latents drift far off the posterior distribution while RDR decodes at matched distance with lower error (Section~\ref{sec:ablations}). A latent-width sweep suggests the advantage grows as the latent widens while the standard objective loses horizon; the trend is descriptive (Section~\ref{sec:bracket}). Preliminary control experiments show RDR more robust to a planner--training rollout mismatch, and a fixed-epoch margin that an optimizer-step-matched control mostly removes (Section~\ref{sec:control}). At matched budget an observation-space predictor reaches the same horizon on this fully observed system; the comparison bounds the latent bottleneck itself, and the RDR contrast is within-latent throughout (Section~\ref{sec:pushforward}).

RDR is a training objective. Symmetry reduction, discrete latents, KL balancing, and latent overshooting all leave a decoder in place with its training distribution unchanged, so RDR applies on top of each; whether the improvement transfers to those settings is untested. Every quantitative claim traces to archived evaluation artifacts produced under a preregistered protocol, and each figure is rendered by a script that re-asserts every quoted number against those artifacts.

\section{Background}
\label{sec:background}

\paragraph{Latent world models.}
During training, observations are available, so latents are \emph{posterior} states anchored to data; multi-step training signals, where present, act in latent space only. At deployment the model \emph{free-runs}: the transition iterates on its own outputs from an encoded initial state, and the decoder is applied to the resulting rollout latents. The RSSM lineage~\cite{hafner2019,hafner2020,hafner2021,hafner2025} trains the decoder on posterior latents throughout; PlaNet~\cite{hafner2019} considered the decode-the-rollout variant (observation overshooting) and judged it too expensive in image domains, and the Dreamer line~\cite{hafner2020,hafner2021,hafner2025} does not revisit it. Decoder-free models (TD-MPC2~\cite{hansen2024} by value prediction, MuDreamer~\cite{burchi2024} by deleting the reconstruction loss) are the one family RDR does not reach, having removed the component it trains.

\paragraph{Valid prediction time.}
Long-horizon forecasts of chaotic systems are scored by VPT: the model free-runs, the decoded forecast's RMSE is normalized against the climatological standard deviation, and VPT is the time at which this normalized error first crosses a threshold, 0.5 by convention, with horizons calibrated in Lyapunov times. For KS at $L{=}22$ we use the literature value $\lambda_{\max} \approx 0.043$~\cite{edson2019}, so one Lyapunov time is 23.26 time units (tu). Published results on this domain (symmetry-reduced manifold models~\cite{linot2020}, full-state reservoir computers~\cite{pathak2018}) are architectural results orthogonal to the objective studied here; Appendix~\ref{app:scale} places every arm on that absolute scale.

\paragraph{Related work.}
The exposure-bias literature addresses train/deployment mismatch at the \emph{input}: scheduled sampling~\cite{bengio2015} mixes model predictions into teacher-forced inputs, and professor forcing~\cite{lamb2016} aligns hidden-state distributions adversarially. Observation-space PDE surrogates (the pushforward trick~\cite{brandstetter2022}, solver-in-the-loop training~\cite{um2020}) unroll the model and train on its own outputs, but they have no latent space and hence no decoder subject to the mismatch. Pixel-space video models (Self Forcing~\cite{huang2025}, Diffusion Forcing~\cite{chen2024}, Next Forcing~\cite{xu2026}) train on their own rollouts without an encoder--latent--decoder triplet. Two concurrent 2026 preprints, NeuroWorld~\cite{dong2026} and Koopman Dreamer~\cite{li2026}, decode rollout states in frozen-dynamics or auxiliary-loss settings; neither isolates the decoder's training distribution as the object of study. Closest is Rollout-LaSDI~\cite{stephany2025}, which trains a latent reduced-order model's decoder on decoded rollout states against reference solutions and cuts maximum error threefold on 2D Burgers. It differs where it matters here: its latent dynamics are per-parameter linear coefficients fit by regression, its system is smooth, it reports no control result, and it holds the latent width fixed; this work studies a learned nonlinear transition iterating on its own output on a chaotic system. Rollout-decoded training is established for latent reduced-order models and, to our knowledge, unmeasured in world models; this work isolates it and measures what it is worth.

\section{Method}
\label{sec:method}

\subsection{Model and base objective}
\label{sec:base}

The model is a deliberately plain encoder--transition--decoder triplet. An MLP encoder $E_\phi$ maps a field snapshot $u_t \in \mathbb{R}^{64}$ to a latent $z_t \in \mathbb{R}^{d}$ (two hidden layers of 256, GELU). A state-space predictor $f_\theta$, a four-layer S5 stack~\cite{smith2023} with state size 64, advances the latent one step. A decoder $D_\psi$, an MLP with one hidden layer of 512, maps latents back to field space. A target encoder $\bar{E}$, the exponential moving average of $E_\phi$ with decay 0.999, provides latent regression targets; we write $\bar{z}_t = \bar{E}(u_t)$ and use \emph{posterior}, by analogy with RSSM practice, for latents computed from observed data.

Four standard terms train the triplet. With $\sg$ the stop-gradient and $\hat{z}_{t+k}$ the free-running rollout ($\hat{z}_t = E_\phi(u_t)$, $\hat{z}_{t+k+1} = f_\theta(\hat{z}_{t+k})$, gradients flowing through the whole chain):
\begin{align*}
L_{\mathrm{TF}} &= \tfrac{1}{T}\textstyle\sum_t \big\| f_\theta(\bar{z}_t) - \sg(\bar{z}_{t+1}) \big\|^2 &&\text{teacher-forced one-step latent prediction,}\\
L_{\mathrm{R}} &= \tfrac{1}{K}\textstyle\sum_{k=1}^{K} \big\| \hat{z}_{t+k} - \sg(\bar{z}_{t+k}) \big\|^2 &&\text{multi-step latent rollout consistency,}\\
L_{\mathrm{OBS}} &= \tfrac{1}{T}\textstyle\sum_t \big\| D_\psi(f_\theta(\bar{z}_t)) - u_{t+1} \big\|^2 &&\text{decode the teacher-forced prediction,}\\
L_{\mathrm{recon}} &= \tfrac{1}{T}\textstyle\sum_t \big\| D_\psi(E_\phi(u_t)) - u_t \big\|^2 &&\text{reconstruction anchor on the online encoder.}
\end{align*}
No term exposes the decoder to a latent more than one teacher-forced step from an observation.

\subsection{The RDR objective}
\label{sec:rdr}

RDR decodes the same free-running rollout that $L_{\mathrm{R}}$ constrains and that evaluation scores, and penalizes its error against the ground-truth fields:
\begin{equation}
L_{\mathrm{RDR}} = \frac{1}{K}\sum_{k=1}^{K} \big\| D_\psi(\hat{z}_{t+k}) - u_{t+k} \big\|^2,
\qquad
L = L_{\mathrm{TF}} + \alpha_e\,L_{\mathrm{R}} + L_{\mathrm{OBS}} + L_{\mathrm{recon}} + \lambda\,L_{\mathrm{RDR}}.
\label{eq:rdr}
\end{equation}
The gradient reaches the decoder directly and, through the rollout chain, the predictor and the encoder, so all three components are shaped by the trajectory the model will actually produce. The curriculum is standard for rollout losses: $\alpha_e$ ramps linearly from 0 to 1 between epochs 2 and 5, the rollout branch (and with it the RDR term) opens after the two warm-up epochs, and the rollout is pure free-running throughout. Setting $\lambda{=}0$ recovers the baseline exactly; we call that arm \emph{posterior-only}. The operating weight is $\lambda{=}0.3$, and the result is flat across $\lambda \in \{0.1, 0.3, 0.6, 1.0\}$ (Section~\ref{sec:lambda}).

The term adds no parameters. Both arms already compute the free-running rollout for $L_{\mathrm{R}}$, so the marginal cost is the $K$ additional decoder evaluations per window and their backward pass, incurred at training time only; inference is unchanged. Appendix~\ref{app:compute} reports step counts, decoder-evaluation counts, wall-clock, and hardware for every arm.

\section{Forecasting Results}
\label{sec:ks}

\subsection{Setup}
\label{sec:setup}

We integrate KS on $L{=}22$ with 64 grid points by ETDRK4 at $\mathrm{d}t{=}0.1$, discarding a 2000-step transient, and generate 512 training, 64 validation, and 64 test trajectories of 256 snapshots each from disjoint seeds. Training draws one fixed 160-snapshot window per trajectory and unrolls the rollout losses over $K{=}128$ steps; batch size 64 (8 optimizer steps per epoch), Adam at $3{\times}10^{-4}$, three seeds per arm. The headline configuration is latent 32, decoder width 512, $\lambda{=}0.3$, 320 epochs.

Evaluation free-runs each model from the first snapshot of every held-out trajectory through the same rollout path used in training, decodes, and scores VPT as Section~\ref{sec:background} defines it. The canonical horizon is 200 steps (20 tu); the long horizon is 1300 steps (130 tu) on separately generated, dealiased records. Appendix~\ref{app:protocol} gives the metric in full, the long-record generation procedure, the pinned-estimator construction behind the canonical/long pair, and a validation-split re-ranking confirming that the reported winner is selectable on validation.

\subsection{Main result}
\label{sec:headline}

\begin{figure}[H]
\centering
\kfig{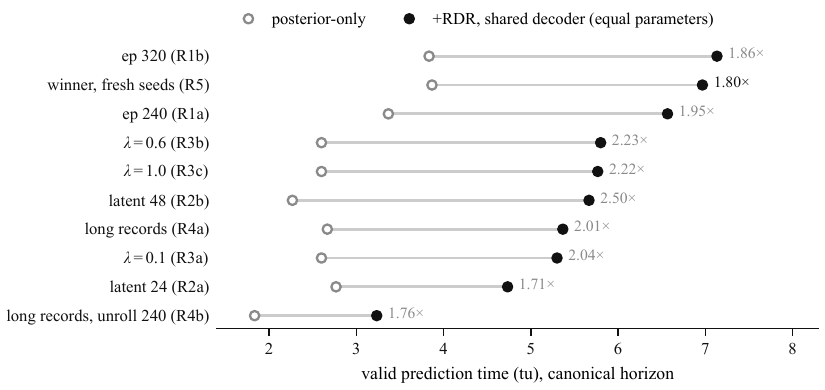}{0.85\linewidth}
\caption{Every preregistered configuration, canonical horizon. Each row is one configuration scored at three seeds; open circles are the posterior-only arm, filled circles the capacity-matched +RDR arm, with the per-row ratio at right. RDR wins 10 of 10, ratio 1.71--2.50$\times$. The fresh-seed confirmation row (R5) is the headline number; rows above it were selection rows.}
\label{fig:forest}
\end{figure}

\begin{figure}[H]
\centering
\kfig{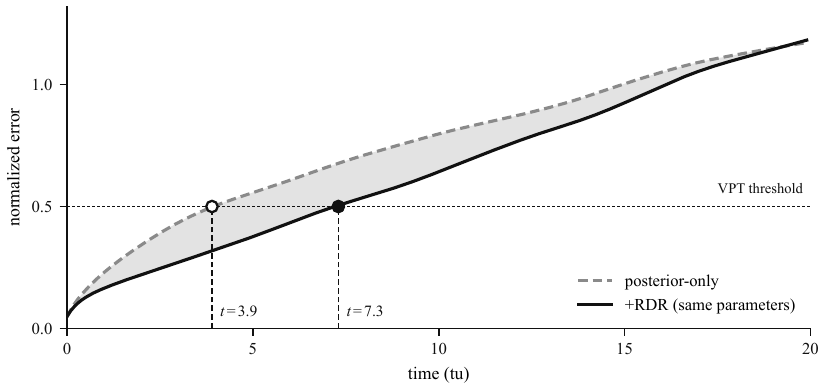}{0.9\linewidth}
\caption{Forecast error against ground truth on one held-out KS trajectory: two world models at an identical 193{,}568 parameters, free-running from the same initial condition. RDR holds lower error at every horizon (shaded); VPT is the first crossing of the 0.5 threshold, $t{=}3.9$ for the posterior-only arm against $t{=}7.3$ for RDR on this trajectory. Arm-level means over all 64 test trajectories at three fresh seeds are $3.87 \pm 0.23$ and $6.97 \pm 0.42$ tu.}
\label{fig:trajectory}
\end{figure}

Table~\ref{tab:sweep} (Appendix~\ref{app:numbers}) and Figure~\ref{fig:forest} summarize the preregistered sweep. Every number in this section is VPT@0.5: the time at which the decoded free-running rollout's normalized RMSE first crosses 0.5, scored over all 64 held-out trajectories and averaged over three seeds per arm (Sections~\ref{sec:background}, \ref{sec:setup}). At fresh seeds (10--12), none used during selection, the posterior-only arm reaches $3.87 \pm 0.23\tu$ and the RDR arm $6.97 \pm 0.42\tu$: a 1.80$\times$ improvement at an identical 193{,}568 trainable parameters (accounting in Appendix~\ref{app:numbers}). The same checkpoints score 3.77 against 6.90 tu on the long horizon, and Figure~\ref{fig:trajectory} shows one held-out trajectory. Across the ten preregistered rows (nine configurations plus the winner's fresh-seed rerun), RDR wins 10 of 10 at capacity-matched ratios of 1.71--2.50$\times$.

\paragraph{The preregistered gate.}
The gate, locked before any sweep number existed, required the preregistered primary arm (variant C, the split-decoder form of RDR defined in Section~\ref{sec:split}) to exceed 5.77 tu, an observation-space baseline measured at a smaller training budget. It does: $6.47 \pm 0.40\tu$ at fresh seeds, every seed above the bar (per-seed 6.4/6.9/6.1); the selection-seed value is $6.90 \pm 0.50$. The bar reproduces at $5.80 \pm 0.44$ on re-measurement (Appendix~\ref{app:protocol}), and Section~\ref{sec:pushforward} gives the same baseline at the winner's budget.

\paragraph{Training-time scaling.}
Training length is the one lever that keeps paying: the primary arm moves from 5.27 to 6.37 to 6.90 tu across 160, 240, and 320 epochs at selection seeds, still unsaturated at the longest budget tested. Longer training is left to future work.

\subsection{Comparison to observation-space prediction}
\label{sec:pushforward}

The strongest non-latent control is an observation-space pushforward predictor~\cite{brandstetter2022} trained on its own unrolled outputs at the winner's budget. It reaches $7.00 \pm 0.26\tu$ (Table~\ref{tab:classes}): parity with the capacity-matched RDR arm, every gap far inside one seed-level sd. Both model classes scale with training at nearly the same rate over the same budget change, the latent stack from 5.27 to 6.90 and observation-space from 5.80 to 7.00. On a fully observed, 64-dimensional system, the latent bottleneck buys nothing over a direct observation-space predictor at matched budget.

This result bounds the latent bottleneck, and it leaves the within-latent RDR contrast untouched. The settings that force a latent (partial observability, pixel observations, a planner that requires a compact state) are the settings in which an observation-space predictor is unavailable; measuring the effect in one of them is the natural next experiment.

\begin{table}[H]
\centering
\caption{Model classes at the winner's training budget (320 epochs), canonical and long horizons, mean $\pm$ sd over three seeds. The pushforward baseline sits between the two latent arms in parameter count. The headline arm matches it within 0.06 combined sd (root sum of squares of the two arms' sds); the preregistered primary arm sits 1.11 combined sd below it.}
\label{tab:classes}
\small
\begin{tabular}{lrcc}
\toprule
model & params & VPT canonical (tu) & VPT long (tu)\\
\midrule
latent + RDR, shared decoder (headline) & 193{,}568 & $6.97 \pm 0.42$ & $6.90 \pm 0.44$\\
latent + RDR, split head (preregistered primary) & 243{,}296 & $6.47 \pm 0.40$ & $6.40 \pm 0.40$\\
observation-space pushforward & 230{,}464 & $7.00 \pm 0.26$ & $6.90 \pm 0.26$\\
latent, posterior-only & 193{,}568 & $3.87 \pm 0.23$ & $3.77 \pm 0.23$\\
\bottomrule
\end{tabular}
\end{table}

\section{Ablations and Diagnostics}
\label{sec:ablations}

\subsection{The split decoder}
\label{sec:split}

Under RDR a single decoder serves two roles that could in principle conflict: sharp reconstruction of posterior latents, and robust decoding of drifted rollout latents. We therefore also evaluate a variant that gives the rollout term its own decode head of identical shape (+25.7\% total parameters), with two evaluation policies: the sharp head everywhere (variant B), or a ramp from sharp to rollout head across the horizon (variant C, fixed by preregistration as the primary arm before any result existed).

The headline arm is the shared-decoder form: one decoder supervised on both posterior and rollout latents, at exactly the posterior-only arm's parameter count. It is the strongest arm. The split variant (243{,}296 parameters against 193{,}568) performs worse in 7 of 10 configurations on both horizons, including the winner and its fresh-seed confirmation (6.47 vs.\ 6.97). The comparison rules out added capacity as the explanation for the gap; it does not by itself establish the mechanism. Variant C is the lower of the two RDR arms, so the gate was cleared by the conservative choice, and its own ratio of 1.67$\times$ is capacity-confounded.

The split's usefulness is latent-size dependent. At latent 128 with a linear decoder the head split is decisive: VPT 0.43 without it, 0.90 with it, under the sharp-head evaluation. At latent 32 with a 512-wide decoder the ordering reverses. The pattern is consistent with a decoder-capacity account: a thin decoder under a wide latent cannot be simultaneously sharp on posterior latents and robust on drifted ones, while a stronger decoder at moderate latent size absorbs both roles and the split only spends parameters.

\subsection{Loss-weight sensitivity}
\label{sec:lambda}

Sweeping $\lambda$ over 0.1, 0.6, and 1.0 against the default 0.3 keeps the capacity-matched ratio between 2.0$\times$ and 2.3$\times$: the term has one hyperparameter and is insensitive to it at this operating point.

\subsection{Latent-width sweep}
\label{sec:bracket}

Re-scoring archived checkpoints under the final protocol completes a four-point latent bracket at one recipe (Figure~\ref{fig:bracket}; full numbers in Appendix~\ref{app:numbers}). The bracket is descriptive: it ran at selection seeds 0--2, post hoc, with no preregistered gate, and the manipulated unit has $n{=}4$ rungs.

\paragraph{The observed pattern.}
As the latent grows from 16 to 48, the posterior-only arm's VPT falls at every rung, from 3.00 tu to 2.77, 2.60, and 2.27, in 3 of 3 seeds at every step. The RDR arm rises through 4.20, 4.73, and 5.90, then holds at 5.67. The pattern is consistent with the mechanism: a wider latent has more directions the decoder never sees under posterior-only training, free-running rollouts wander into exactly those directions, and RDR trains the decoder there. RDR wins at every rung and in every seed; all 36 per-seed paired ratios exceed 1, with a minimum of 1.32 on the shared arm.

\begin{figure}[H]
\centering
\kfig{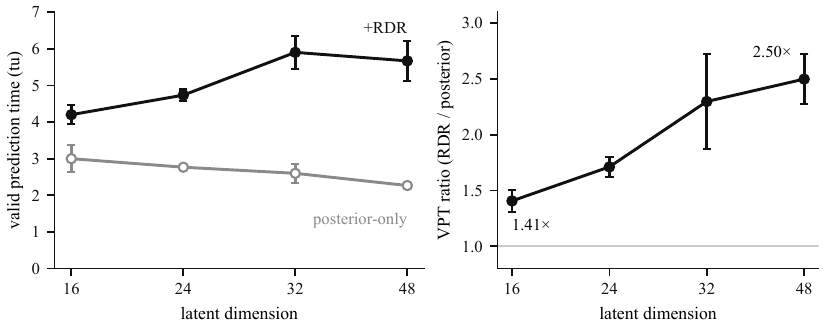}{0.9\linewidth}
\caption{The latent-size bracket at the matched recipe (decoder 512, 160 epochs, $\lambda{=}0.3$, final protocol, three paired seeds per rung). Left: VPT by arm; the posterior-only arm falls as the latent widens while the RDR arm rises, its top step within noise. Right: the per-rung paired ratio rises from 1.41$\times$ to 2.50$\times$; the endpoints are separated by roughly 3.4 combined sd (sum of the endpoint sds), and the top step holds in 2 of 3 seeds. The bracket is descriptive: selection seeds, post hoc, no preregistered gate. Error bars are $\pm 1$ sd over seeds.}
\label{fig:bracket}
\end{figure}

\paragraph{Statistical strength.}
The ratio increases across the bracket: the 16-to-24 and 24-to-32 steps hold in 3 of 3 paired seeds, and the endpoints ($1.41 \pm 0.10$ vs.\ $2.50 \pm 0.22$) are cleanly separated. The top step is noisy: 32 to 48 holds in 2 of 3 seeds and reverses under leave-one-seed-out. The manipulated unit is the rung, with $n{=}4$; the observed ordering attains the smallest cluster-permutation $p$ available, $2/24 \approx 0.083$, and a per-seed Spearman correlation (0.907 over 12 points) would overstate certainty if quoted alone. Predictor width covaries with latent size by construction, so the axis varies latent and predictor width together; the per-rung arm contrast is unaffected, since both arms share the width. The posterior arm falls below the persistence baseline (2.40 tu) at latent 48 in 3 of 3 seeds, so baseline collapse contributes part of the top-rung ratio growth. VPT is quantized at 0.1 tu, so the smallest reported seed spreads are quantization-limited. Long-horizon ratios track the canonical values throughout (1.40/1.74/2.30/2.46).

The rung nearest the system's inertial manifold (dimension near 8 for $L{=}22$ KS) is where RDR's edge is weakest, and the edge grows away from it. Latent ${\approx}8$ was not run; behavior below 16 is untested, and nothing in $[16, 48]$ suggests a turnover.

\subsection{Diagnostics of the latent mismatch}
\label{sec:depth}

\begin{figure}[H]
\centering
\kfig{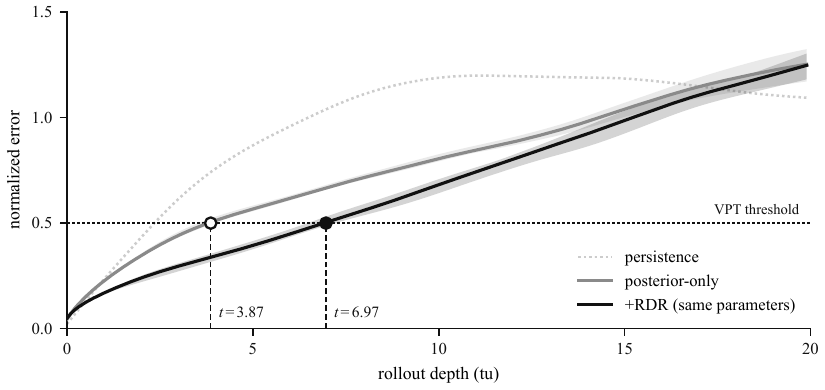}{0.85\linewidth}
\caption{Normalized error of the decoded free-running rollout against rollout depth, mean over the 64 held-out test trajectories, three fresh seeds per arm (seed spread shaded). The posterior-only arm crosses the 0.5 threshold at $3.87\tu$ on average and the RDR arm at $6.97\tu$. Each seed's crossing of its own curve is that seed's reported VPT, so this figure and the headline number are one artifact. The persistence forecast is the dotted reference.}
\label{fig:depth}
\end{figure}

\paragraph{Decoded error against rollout depth.}
Figure~\ref{fig:depth} plots the normalized error of the decoded free-running rollout against rollout depth, averaged over all 64 held-out trajectories at the fresh seeds. Both arms decode the early rollout accurately, where the latents remain close to the posterior states the standard objective trains on. The posterior-only arm then crosses the 0.5 threshold at $3.87\tu$ while the RDR arm holds below it until $6.97\tu$, and the separation persists far past both crossings, closing only where both errors approach the climatological scale. The divergence begins at the depths teacher forcing never supervises, exactly where the two objectives differ.

\begin{figure}[H]
\centering
\kfig{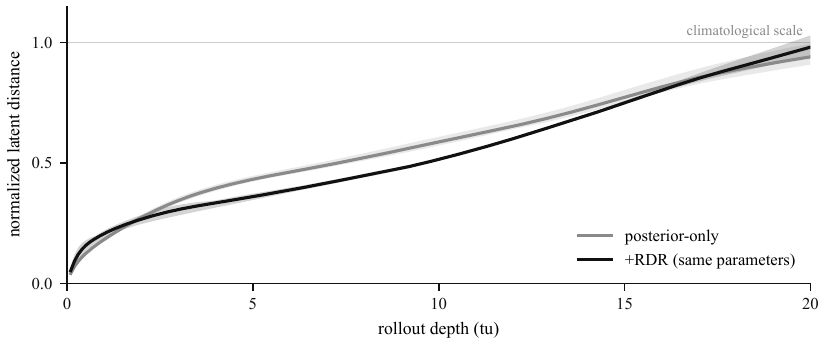}{0.85\linewidth}
\caption{Distance between the free-running rollout latent and the posterior latent of the true state at the same time, against rollout depth: mean over the 64 held-out test trajectories, three fresh seeds per arm (seed spread shaded), normalized per model by its climatological latent scale so the two arms' latent spaces are comparable. Both arms drift far off the posterior distribution and saturate near the climatological scale.}
\label{fig:mismatch}
\end{figure}

\paragraph{Distance between posterior and free-running latents.}
Figure~\ref{fig:mismatch} measures the mismatch itself: the distance between the rollout latent and the posterior latent of the true state at the same time, computed from the same rollouts through the same encode-and-roll path evaluation uses. Distances are normalized per model by its climatological latent scale (the all-pairs mean distance between posterior latents of unrelated states), so a value of 1 means the rollout latent is as far from the true state's latent as two unrelated states are from each other. Both arms leave the posterior distribution: by a depth of $3.9\tu$ the normalized distance reaches 0.39 for the posterior-only arm and 0.33 for RDR, and both saturate near 1 by the end of the window. Two observations follow. RDR's latents drift somewhat less at mid-horizon, so improved latent dynamics contribute to the gain. And at matched normalized distance RDR decodes with lower field error (posterior 0.38 against RDR 0.28 at distance 0.3, 0.52 against 0.47 at 0.4; 3 of 3 seeds at every sampled distance), so the decoder itself handles off-distribution latents better. The diagnostic shows both pathways active and does not apportion the gain between them (Section~\ref{sec:limitations}).

\section{Control Experiments}
\label{sec:control}

We carry the objective to two classic control tasks, pendulum swing-up and a cartpole swing-up variant, with action-conditioned world models: ensembles of ten models per arm drive a cross-entropy-method MPC planner~\cite{chua2018,pinneri2020} with 128 sampled action sequences per step and no disagreement bonus, scored over 20 paired episodes with identical episode seeds for both arms. These results are preliminary: the largest margins in this section arise under a protocol that confounds data volume with optimizer-step count, and a step-matched control removes most of them.

\subsection{Step-efficiency}
\label{sec:ladder}

We vary the training-data budget under two protocols. Holding training epochs fixed while the dataset shrinks also shrinks the number of optimizer updates (480/240/60 steps at the 50/25/10\% rungs against the full-data 960), so a fixed-epoch ladder varies data volume and optimization length together. We therefore run the ladder twice: at fixed epochs, and with optimizer steps matched at every rung.

\begin{figure}[H]
\centering
\kfig{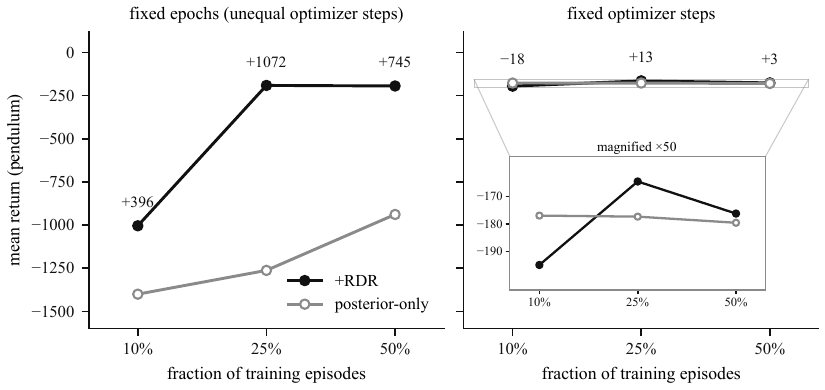}{0.95\linewidth}
\caption{The reduced-data ladder on pendulum, mean return over 20 paired episodes, annotations giving the RDR-minus-posterior margin at each rung. Left: epochs held fixed, so reduced rungs receive proportionally fewer optimizer steps; RDR wins 20/20 paired episodes at every rung, holds near its full-data return down to the 25\% rung, and the posterior arm degrades at every rung. Right: optimizer steps matched at every rung; the margins narrow to $+3$/$+13$/$-18$, the sign reverses at two cartpole rungs (6/20 paired wins at 25\% and 50\%), and at the shared scale the two arms coincide. The inset magnifies the marked sliver, across which the entire spread is about 30 return.}
\label{fig:ladder}
\end{figure}

At fixed epochs, RDR wins 20 of 20 paired episodes at every reduced rung on both tasks, with pendulum mean-return margins of +745, +1072, and +396 (Figure~\ref{fig:ladder}, left). With steps matched, both arms train fully at every rung (the pendulum posterior arm at 25\% exceeds its own full-data score) and the margins narrow to $+3$/$+13$/$-18$, the sign reversing at two cartpole rungs (Figure~\ref{fig:ladder}, right). Pendulum returns at every reduced rung sit close to the full-data level; on cartpole both arms fall below their full-data scores together (Table~\ref{tab:ladder}). The pair locates the benefit: RDR reaches useful control in fewer optimizer steps, and with steps matched most of the advantage disappears. We make no sample-efficiency claim.

\subsection{Robustness to planner--training rollout mismatch}
\label{sec:mismatch}

A deployed planner can roll the world model with an implementation that differs from the one used in training; a common variant re-initializes the recurrent state at each planning step. We evaluate both arms under the stateful rollout, which matches training, and under the per-step-reset variant, on both tasks and both seed sets. Moving to the stateful planner improves both arms in all four rows, and the posterior arm gains more in each: +9.0, +18.1, +17.1, and +12.7 return against RDR's +1.8, +3.6, +12.8, and +10.6 (Table~\ref{tab:planner}). RDR is more robust to planner--training rollout mismatch than posterior-only training, and the asymmetry is consistent with the objective: a decoder trained on free-running rollouts loses less when the planner's rollout function changes. Under the stateful planner the arms perform equivalently (margins +0.61 and +1.24 on the primary rows), which bounds the property: RDR reduces the cost of the mismatch and offers no additional planning advantage once it is removed.

\paragraph{Scope of the oracle comparison.}
Under the stateful planner, ensemble MPC over the learned world models matches planning with the true dynamics at the same search budget: on pendulum the RDR ensemble returns $-177.45$ against the oracle planner's $-183.67$, on cartpole 158.57 against 151.01. The posterior-only ensemble shares the property ($-178.06$ and 157.33 on the same rows), so oracle-level planning is a property of the ensembles and carries no RDR contrast.

\section{Limitations and Conclusion}
\label{sec:limitations}

Four limitations bound the claims.

\emph{Benchmark scope.} The forecasting claim rests on one dynamical system at one domain size, three seeds per cell, and a metric quantized at 0.1 tu; the uniform 10-of-10 direction and the fresh-seed confirmation mitigate this without removing it. The control results cover two low-dimensional classic tasks and are preliminary. Transfer to other systems, backbones, and observation modalities is untested.

\emph{Training cost.} RDR adds training-time compute: +40\% decoder evaluations and a 5--10\% wall-clock overhead per run at the headline configuration (Appendix~\ref{app:compute}).

\emph{Mechanism isolation.} The split-decoder ablation rules out added capacity, and the diagnostics of Section~\ref{sec:depth} show both pathways active. The pathway itself is not isolated: the RDR gradient reaches the decoder, the transition, and the encoder, and these experiments do not apportion the gain among decoder robustness, improved latent dynamics, and implicit regularization.

\emph{Selection and scaling.} Configuration selection during the sweep compared results on the test split; validation-split re-ranking selects the same winner with rank agreement $\rho = 0.979$ (Appendix~\ref{app:protocol}), and the headline is reported at fresh seeds never used in selection. The latent-width trend is descriptive (Section~\ref{sec:bracket}), and on this system the latent stack is not shown to beat observation-space training (Section~\ref{sec:pushforward}).

What is shown: a latent world model's decoder receives almost no training signal on the free-running rollout it must decode at deployment, and Rollout-Decoded Reconstruction closes that gap with one loss term and no new parameters. On a chaotic PDE the term raises valid prediction time by 1.80$\times$ in 10 of 10 preregistered configurations, and wherever a decoder is retained it is available at the cost of one weight and the training-time compute of Appendix~\ref{app:compute}.

What is not yet shown is generality. The remaining validation runs through the regimes that force a latent (partial observability, pixel observations, planning over a compact state), additional systems and backbones, and the gradient-pathway ablations that would isolate why the term works. We leave that to follow-up work.

\appendix

\section{Evaluation Protocol Details}
\label{app:protocol}

\paragraph{Metric.}
For each arm, the model free-runs from the first snapshot of each of the 64 held-out trajectories; the decoded rollout is scored by RMSE over (trajectory, space) at each horizon step, normalized by the climatological standard deviation of the scored window. VPT@0.5 is the time at which this normalized error first crosses 0.5, reported in time units on the 0.1-tu sampling grid; a curve that never crosses is right-censored at the horizon. Canonical horizon: 200 steps on the held-out test split. Long horizon: 1300 steps on 64 separately generated long records. The split-head ramp evaluation blends its two heads with a denominator pinned to 200 steps regardless of the evaluated horizon, so the canonical and long evaluations of one checkpoint are a single estimator scored at two horizons; the canonical/long pair is a censoring comparison on one pinned forecaster.

\paragraph{Long-record generation.}
Long-horizon ground truth is generated with 2/3-rule dealiasing and a per-step Hermitian projection.

\paragraph{Validation-split re-ranking.}
All sweep configurations were re-scored on the validation split (120 evaluations, no retraining; no trainer ever selected weights on validation). Validation selects the same winner, rank agreement with the test ranking is Spearman $\rho = 0.979$ over the nine configurations (0.985 with the fresh-seed rerun included), and the epoch-scaling verdict holds on validation (6.73 at 240 epochs rising to 7.13 at 320, the top two). The reported test numbers are held-out estimates for a validation-selectable configuration.

\paragraph{The 5.77 bar.}
The preregistered gate's threshold of 5.77 tu re-measures at $5.80 \pm 0.44\tu$ under the final protocol, a 0.03 tu match; at the winner's budget the same observation-space baseline reaches $7.00 \pm 0.26$ (Section~\ref{sec:pushforward}).

\section{Full Numerical Results}
\label{app:numbers}

\begin{table}[H]
\centering
\caption{The preregistered sweep, all four arms, canonical horizon, VPT mean $\pm$ sd (tu) over three seeds (sd suppressed where the archived summary reports arm means only). Long-horizon columns track canonical throughout (every arm within 0.15 tu of its canonical value) and are omitted for width; the winner's long-horizon values appear in Section~\ref{sec:headline}. Rows sorted by shared-arm VPT. R5 is the winner configuration at fresh seeds 10--12; all other rows are selection rows at seeds 0--2.}
\label{tab:sweep}
\footnotesize
\begin{tabular}{lcccc}
\toprule
config & posterior & +RDR shared & +RDR split, sharp (B) & +RDR split, ramp (C)\\
\midrule
R1b: 320 epochs & $3.83 \pm 0.21$ & $7.13 \pm 0.50$ & 6.53 & $6.90 \pm 0.50$\\
R5: winner, fresh seeds & $3.87 \pm 0.23$ & $6.97 \pm 0.42$ & 6.13 & $6.47 \pm 0.40$\\
R1a: 240 epochs & 3.37 & $6.57 \pm 0.71$ & 6.10 & $6.37 \pm 0.67$\\
R3b: $\lambda{=}0.6$ & 2.60 & $5.80 \pm 0.70$ & 5.10 & $5.57 \pm 0.76$\\
R3c: $\lambda{=}1.0$ & 2.60 & $5.77 \pm 0.15$ & 5.43 & $6.07 \pm 0.71$\\
R2b: latent 48 & 2.27 & $5.67 \pm 0.55$ & 5.47 & $5.80 \pm 1.35$\\
R4a: long records & 2.67 & $5.37 \pm 0.12$ & 4.57 & $4.73 \pm 0.29$\\
R3a: $\lambda{=}0.1$ & 2.60 & $5.30 \pm 0.66$ & 4.40 & $4.47 \pm 0.40$\\
R2a: latent 24 & 2.77 & $4.73 \pm 0.15$ & 4.53 & $4.73 \pm 0.40$\\
R4b: long records, unroll 240 & 1.83 & $3.23 \pm 0.64$ & 2.27 & $2.37 \pm 0.15$\\
\bottomrule
\end{tabular}
\end{table}

The window dataset draws one fixed 160-of-256-snapshot window per trajectory for the entire run, so 37.5\% of each training record is never seen at any epoch; this is why the longer-record rows (R4a, R4b) do not improve on their shorter-record counterparts.

\begin{table}[H]
\centering
\caption{Per-seed VPT (tu), canonical horizon, for the two parameter-matched arms of every preregistered configuration. Every one of the 30 paired comparisons favors RDR. Per-seed values for the split-decoder arms and the long horizon are in the archived evaluation artifacts.}
\label{tab:perseed}
\footnotesize
\begin{tabular}{lccc}
\toprule
config & seeds & posterior per-seed & +RDR shared per-seed\\
\midrule
R1a: 240 epochs & 0/1/2 & 3.4 / 2.9 / 3.8 & 6.7 / 7.2 / 5.8\\
R1b: 320 epochs & 0/1/2 & 4.0 / 3.6 / 3.9 & 7.2 / 7.6 / 6.6\\
R2a: latent 24 & 0/1/2 & 2.7 / 2.8 / 2.8 & 4.9 / 4.7 / 4.6\\
R2b: latent 48 & 0/1/2 & 2.3 / 2.2 / 2.3 & 5.3 / 5.4 / 6.3\\
R3a: $\lambda{=}0.1$ & 0/1/2 & 2.7 / 2.3 / 2.8 & 5.9 / 5.4 / 4.6\\
R3b: $\lambda{=}0.6$ & 0/1/2 & 2.7 / 2.3 / 2.8 & 6.5 / 5.8 / 5.1\\
R3c: $\lambda{=}1.0$ & 0/1/2 & 2.7 / 2.3 / 2.8 & 5.9 / 5.8 / 5.6\\
R4a: long records & 0/1/2 & 2.7 / 2.4 / 2.9 & 5.3 / 5.3 / 5.5\\
R4b: long records, unroll 240 & 0/1/2 & 1.8 / 1.9 / 1.8 & 2.5 / 3.6 / 3.6\\
R5: winner, fresh seeds & 10/11/12 & 4.0 / 3.6 / 4.0 & 7.3 / 7.1 / 6.5\\
\bottomrule
\end{tabular}
\end{table}

\begin{table}[H]
\centering
\caption{The latent-size bracket (Figure~\ref{fig:bracket}), canonical horizon, matched recipe (decoder 512, 160 epochs, $\lambda{=}0.3$), three paired seeds per rung. Ratios are per-seed paired means. All four rungs are scored under the final protocol at the same recipe.}
\label{tab:bracket}
\small
\begin{tabular}{ccccc}
\toprule
latent & posterior & +RDR shared & +RDR split, ramp (C) & ratio shared/posterior\\
\midrule
16 & $3.00 \pm 0.36$ & $4.20 \pm 0.27$ & $4.03 \pm 0.21$ & $1.41 \pm 0.10$\\
24 & $2.77 \pm 0.06$ & $4.73 \pm 0.15$ & $4.73 \pm 0.40$ & $1.71 \pm 0.09$\\
32 & $2.60 \pm 0.27$ & $5.90 \pm 0.46$ & $5.27 \pm 0.40$ & $2.30 \pm 0.43$\\
48 & $2.27 \pm 0.06$ & $5.67 \pm 0.55$ & $5.80 \pm 1.35$ & $2.50 \pm 0.22$\\
\bottomrule
\end{tabular}
\end{table}

\begin{table}[H]
\centering
\caption{The reduced-data ladder, both regimes, both tasks: mean return of each arm with the RDR-minus-posterior paired margin and RDR's paired win count over 20 episodes. Fixed-epoch rungs receive 480/240/60 optimizer steps as the data shrinks; step-matched rungs all receive the full-data 960 and were evaluated under the stateful planner, whose full-data rows appear in Table~\ref{tab:planner}. Rungs draw episode windows independently, so smaller rungs are not strict subsets of larger ones. Margins are paired, computed from unrounded per-episode returns, and can differ in the last digit from the difference of the rounded arm means.}
\label{tab:ladder}
\small
\begin{tabular}{llcccc}
\toprule
regime & data & pend.\ post / +RDR & $\Delta$ (wins) & cartp.\ post / +RDR & $\Delta$ (wins)\\
\midrule
fixed epochs & 10\% & $-1400$ / $-1004$ & $+396$ (20/20) & $-28$ / $+21$ & $+48$ (20/20)\\
 & 25\% & $-1263$ / $-191$ & $+1072$ (20/20) & $-21$ / $+110$ & $+132$ (20/20)\\
 & 50\% & $-939$ / $-194$ & $+745$ (20/20) & $+49$ / $+114$ & $+64$ (20/20)\\
\midrule
fixed steps & 10\% & $-177.0$ / $-194.9$ & $-18.0$ (12/20) & 90.3 / 95.6 & $+5.4$ (17/20)\\
 & 25\% & $-177.4$ / $-164.6$ & $+12.8$ (12/20) & 122.9 / 115.6 & $-7.4$ (6/20)\\
 & 50\% & $-179.6$ / $-176.2$ & $+3.3$ (18/20) & 129.2 / 124.8 & $-4.3$ (6/20)\\
\bottomrule
\end{tabular}
\end{table}

\begin{table}[H]
\centering
\caption{The planner rollout comparison, all four measured rows: mean return per arm under the per-step-reset and stateful planner rollouts, with the paired margin and RDR's win count over 20 episodes. The stateful rollout matches the one used in training. The last two columns give each arm's gain from moving to the stateful planner: the posterior arm gains more in 4 of 4 rows. Margins are computed as in Table~\ref{tab:ladder}.}
\label{tab:planner}
\footnotesize
\begin{tabular}{llccccc}
\toprule
row & planner & posterior & +RDR & $\Delta$ (wins) & post.\ gain & RDR gain\\
\midrule
pendulum, seeds 0--19 & per-step-reset & $-187.10$ & $-179.27$ & $+7.83$ (19/20) & & \\
 & stateful & $-178.06$ & $-177.45$ & $+0.61$ (9/20) & $+9.04$ & $+1.82$\\
pendulum, seeds 100--119 & per-step-reset & $-263.02$ & $-247.28$ & $+15.74$ (18/20) & & \\
 & stateful & $-244.90$ & $-243.69$ & $+1.22$ (13/20) & $+18.12$ & $+3.59$\\
cartpole, seeds 0--19 & per-step-reset & 140.25 & 145.77 & $+5.52$ (15/20) & & \\
 & stateful & 157.33 & 158.57 & $+1.24$ (11/20) & $+17.08$ & $+12.80$\\
cartpole, seeds 100--119 & per-step-reset & 141.32 & 142.81 & $+1.49$ (17/20) & & \\
 & stateful & 154.03 & 153.45 & $-0.59$ (13/20) & $+12.71$ & $+10.64$\\
\bottomrule
\end{tabular}
\end{table}

\paragraph{Parameter accounting.}
All counts exclude the 90{,}656-parameter EMA target encoder, a non-gradient shadow copy unused at evaluation, from every arm identically (full checkpoint totals: 284{,}224 shared/posterior, 333{,}952 split). The observation-space pushforward's 230{,}464 sits between the two latent arms' counts, so the class comparison of Table~\ref{tab:classes} is not a capacity artifact in either direction.

\section{Compute Accounting}
\label{app:compute}

This appendix reports the training cost RDR adds. All numbers are for the headline configuration (latent 32, decoder 512, window 160, $K{=}128$, batch 64, 320 epochs) and come from the training logs and artifact timestamps of the archived runs.

\begin{table}[H]
\centering
\caption{Compute accounting at the headline configuration, per training run on one NVIDIA H100. A decoder evaluation is one application of $D_\psi$ to one latent vector. Both arms compute the $K{=}128$ free-running rollout for $L_{\mathrm{R}}$; RDR additionally decodes each rollout latent, which is its entire marginal cost. Inference is identical for both arms.}
\label{tab:compute}
\small
\begin{tabular}{lcc}
\toprule
quantity & posterior-only & +RDR (shared)\\
\midrule
trainable parameters & 193{,}568 & 193{,}568\\
batch size & 64 & 64\\
optimizer steps (8 per epoch) & 2{,}560 & 2{,}560\\
free-running rollout length $K$ & 128 & 128\\
decoder evaluations per optimizer step & 20{,}416 & 28{,}608\\
decoder evaluations per training run & $52.3$M & $73.1$M\\
training FLOPs per run (estimated) & $5.0 \times 10^{13}$ & $5.6 \times 10^{13}$\\
wall-clock per training run & 27--28 min & 28--30 min\\
\bottomrule
\end{tabular}
\end{table}

Per optimizer step, both arms decode the teacher-forced predictions ($64 \times 159$ latents) and the reconstruction anchor ($64 \times 160$); RDR adds one decode of each rollout latent ($64 \times 128$ per step), a +40\% increase in decoder evaluations. The measured wall-clock overhead is 5--10\% per run. The rollout branch opens after the two warm-up epochs, so the RDR term is active for 2{,}544 of the 2{,}560 steps. FLOPs are estimated by parameter counting from the checkpoint component sizes (encoder 90{,}656, predictor 53{,}184, decoder 49{,}728): two FLOPs per multiply-accumulate forward, backward at twice forward, and the EMA target encoder forward-only. The per-step estimate is $1.94 \times 10^{10}$ against $2.18 \times 10^{10}$, a +12.6\% difference on which the measured 5--10\% wall-clock overhead sits.

The preregistered sweep comprised 90 training runs and 210 evaluations, roughly 29 H100-hours (training plus evaluation) on a short-lived 4$\times$H100 spot instance; total GPU spend for the program was roughly \$155, including one spot-preempted attempt whose numbers were discarded. Every evaluation and figure in this paper runs on a laptop CPU from the archived checkpoints. Peak GPU memory was not instrumented.

\section{The Absolute Scale}
\label{app:scale}

This appendix places every arm on the absolute VPT scale, so the within-architecture effects reported above can be read against published results on this domain. The classes differ in structure and evaluation protocol; the caveats after the table govern any cross-class reading.

\begin{table}[H]
\centering
\caption{Every arm on the absolute VPT scale for KS at $L{=}22$, with $\lambda_{\max}{=}0.043$~\cite{edson2019} (one Lyapunov time = 23.26 tu). The published band and the reservoir row are different settings and carry the caveats below; neither is class-comparable to the latent arms.}
\label{tab:scale}
\small
\begin{tabular}{lccl}
\toprule
tier & VPT (tu) & $\Lambda t$ & status\\
\midrule
persistence baseline & 2.4 & 0.10 & \\
latent, posterior-only (winner config) & 3.87 & 0.17 & measured, fresh seeds\\
observation-space pushforward, small budget & 5.77 & 0.25 & the preregistered gate's bar\\
latent + RDR, split head (primary arm) & 6.47 & 0.28 & fresh-seed confirmed\\
latent + RDR, shared decoder (headline) & 6.97 & 0.30 & fresh-seed confirmed\\
observation-space pushforward, matched budget & 7.00 & 0.30 & Section~\ref{sec:pushforward}\\
published latent-ROM band~\cite{linot2020} & 33--47 & 1.4--2.0 & symmetry-reduced\\
full-state reservoir (ESN), this work & 73.0 & 3.14 & see caveats\\
\bottomrule
\end{tabular}
\end{table}

The published band uses symmetry reduction (the slice trick), an architectural technique our models do not employ. It is orthogonal to the objective studied here: that model is itself an encoder--dynamics--decoder triplet whose decoder is fit by reconstruction, so RDR composes with it. The ESN, implemented to calibrate what this data and metric support, is full-state with no latent bottleneck, receives a 50-snapshot teacher-forced spin-up that the latent arms do not (their forecasts cold-start from a single snapshot), and is retuned per horizon; the canonical-horizon tuning right-censors at the 20-tu window, and the quoted 73.0 comes from the long-horizon tuning. It demonstrates that the data and metric support far longer horizons than any bottlenecked arm here reaches, and it is never class-compared against them.

\end{document}